\documentclass[manuscript,screen]{acmart}
\renewcommand\footnotetextcopyrightpermission[1]{}
\AtBeginDocument{%
  \providecommand\BibTeX{{%
    \normalfont B\kern-0.5em{\scshape i\kern-0.25em b}\kern-0.8em\TeX}}}

\usepackage{multirow}

\let\vec\boldsymbol

\newcommand{\R}{\mathbb{R}}
\newcommand{\E}{\mathbb{E}}

\usepackage{url}
\usepackage{makecell,rotating}

\setcopyright{acmcopyright}
\copyrightyear{2018}
\acmYear{2018}
\acmDOI{10.1145/1122445.1122456}

\begin{document}

\title{DiMBERT: Learning Vision-Language Grounded Representations with Disentangled Multimodal-Attention}

\author{Fenglin Liu}
\affiliation{\institution{ADSPLAB, School of ECE, Peking University}}
\email{fenglinliu98@pku.edu.cn}

\author{Xian Wu}
\affiliation{\institution{Tencent}}
\email{kevinxwu@tencent.com}

\author{Shen Ge}
\affiliation{\institution{Tencent}}
\email{shenge@tencent.com}

\author{Xuancheng Ren}
\affiliation{\institution{MOE Key Laboratory of Computational Linguistics, School of EECS, Peking University}}
\email{renxc@pku.edu.cn}

\author{Wei Fan}
\affiliation{\institution{Tencent}}
\email{Davidwfan@tencent.com}

\author{Xu Sun}
\affiliation{\institution{School of EECS, Peking University}}
\affiliation{\institution{Center for Data Science, Peking University}}
\email{xusun@pku.edu.cn}

\author{Yuexian Zou}
\affiliation{\institution{ADSPLAB, School of ECE, Peking University}}
\email{zouyx@pku.edu.cn}

\renewcommand{\shortauthors}{Liu et al.}
\begin{abstract}
Vision-and-language (V-L) tasks require the system to understand both vision content and natural language, thus learning fine-grained joint representations of vision and language (a.k.a. V-L representations) are of paramount importance. Recently, various pre-trained V-L models are proposed to learn V-L representations and achieve improved results in many tasks.
However, the mainstream models process both vision and language inputs with the same set of attention matrices. As a result, the generated V-L representations are {\it entangled} in \textit{one common latent space}. To tackle this problem, we propose DiMBERT (short for \textbf{Di}sentangled \textbf{M}ultimodal-Attention \textbf{BERT}), which is a novel framework that applies separated attention spaces for vision and language,
and the representations of multi-modalities can thus be disentangled explicitly. To enhance the correlation between vision and language in disentangled spaces, we introduce the visual concepts to DiMBERT which represent visual information in textual format. In this manner, visual concepts help to bridge the gap between the two modalities.
We pre-train DiMBERT on a large amount of image-sentence pairs on two tasks: bidirectional language modeling and sequence-to-sequence language modeling.
After pre-train, DiMBERT is further fine-tuned for the downstream tasks.
Experiments show that DiMBERT sets new state-of-the-art performance on three tasks (over four datasets), including both generation tasks (image captioning and visual storytelling) and classification tasks (referring expressions). The proposed DiM (short for \textbf{Di}sentangled \textbf{M}ultimodal-Attention) module can be easily incorporated into existing pre-trained V-L models to boost their performance, up to a 5\% increase on the representative task. Finally, we conduct a systematic analysis and demonstrate the effectiveness of our DiM and the introduced visual concepts.
\end{abstract}



\maketitle

\begin{table*}[t]
\footnotesize
\setlength{\tabcolsep}{1.2pt}
\caption{Comparison between our DiMBERT and other works on learning vision-language representations. The \textbf{\color{red} Red} colored texts indicate differences from most existing works. ESA and DiM stands for the Entangled Self-Attention and Disentangled Multimodal-Attention. ISRP, BLM, MOP, S2SLM are short for Image-Sentence Relationship Prediction, Bidirectional Language Modeling, Masked Object Prediction and Seq-to-Seq Language Modeling, respectively.}
\label{tab:compare}
\begin{tabular}{@{}l|c|c|c|c|l|c@{}}
\toprule

Method  & Basic Module & Visual Features  & Textual Features & Pre-train Captioning Datasets  & Pre-training Tasks  & Downstream Tasks   \\ \midrule[\heavyrulewidth]

B2T2 \cite{Alberti2019B2T2}  & \multirow{12}{*}{\begin{tabular}[c]{@{}c@{}} ESA-based \\ Transformer \end{tabular}} & \multirow{12}{*}{Image RoIs}  & \multirow{12}{*}{Sentence Words} & Conceptual Captions \cite{Sharma2018conceptual} & ISRP + BLM & \multirow{12}{*}{\begin{tabular}[c]{@{}c@{}} Classification Task \end{tabular}}  \\ 
\cmidrule(){1-1} \cmidrule(){5-6}

VisualBERT \cite{Li2019VisualBERT}  &  &   & & MSCOCO caption \cite{chen2015microsoft}  & ISRP + BLM &     \\ 
\cmidrule(){1-1} \cmidrule(){5-6}

Unicoder-VL \cite{Li2019unicoder-VL}  &  &   & & Conceptual Captions \cite{Sharma2018conceptual} & ISRP + BLM + MOP  &   \\
\cmidrule(){1-1} \cmidrule(){5-6}

VL-BERT \cite{Su2019vlbert}  & & &  & Conceptual Captions \cite{Sharma2018conceptual} & BLM + MOP  &    \\
\cmidrule(){1-1} \cmidrule(){5-6}

UNITER \cite{chen2019UNITER}  &  & &  & \begin{tabular}[c]{@{}c@{}} Conceptual Captions \cite{Sharma2018conceptual} \\ + VG Captions \cite{krishna2017visualgenome} \\ + MSCOCO caption \cite{chen2015microsoft} \\ + SBU Captions \cite{Ordonez2011Im2Text} \end{tabular} & ISRP + BLM + MOP   &  \\ 
\cmidrule(){1-1} \cmidrule(){5-7}

VLP \cite{Zhou2019VLP}  &  & &  &  Conceptual Captions \cite{Sharma2018conceptual}  & BLM + S2SLM & \multirow{4}{*}{\begin{tabular}[c]{@{}c@{}} \ \ Classification Task \\ + \textbf{\color{red} Generation Task} \end{tabular}} \\ 
\cmidrule(){1-6}

DiMBER [\textbf{\color{red}Ours}]  & \begin{tabular}[c]{@{}c@{}} \textbf{\color{red} DiM}-based \\ Transformer \end{tabular}  &\begin{tabular}[c]{@{}c@{}}  Image RoIs \\ + \textbf{\color{red} Visual Concepts} \end{tabular} & Sentence Words & \begin{tabular}[c]{@{}c@{}} Conceptual Captions \cite{Sharma2018conceptual}  \end{tabular}  & BLM + S2SLM  &  \\ 

\bottomrule
\end{tabular}
\end{table*}

\section{Introduction} \label{sec:intro}
Recently, there is a surge of research interests in vision-and-language (V-L) tasks, such as image captioning \cite{chen2015microsoft} and visual storytelling \cite{visualstorytelling}. In V-L tasks, it is vital to learn the alignments and relationships between V-L modalities and generate fine-grained V-L representations \cite{lu2019vilbert,Su2019vlbert,chen2019UNITER,liu2019MIA}.
However, many existing systems are task-specific models, focusing on individual tasks only.
As a result, learning universal V-L representations and empowering models with the ability to adapt to a wide range of downstream V-L tasks are now becoming the critical topics in current V-L research.

Inspired by the successful pre-training models like ResNet \cite{he2016deep} in computer vision and BERT \cite{devlin2018bert} in natural language processing, several attempts \cite{lu2019vilbert,Tan2019LXMERT,Alberti2019B2T2,Zhou2019VLP,Su2019vlbert} have been conducted to learn such universal V-L representations. 
Table~\ref{tab:compare} summarizes some representative works in image domain. 
As we can see, most systems adopt the Transformer framework \cite{devlin2018bert,ashish2017attention} as the backbone, feeding both visual and textual features into the same stacked transformers.
In the pre-train stage, these systems use large-scale image-sentence pairs to adapt the transformers to the V-L scenarios; In the fine-tune stage, the transformers are further optimized for downstream tasks. Although these pre-training V-L systems receive performance gains across multiple downstream tasks, there are still some potential directions for further improvement:

    $\bullet$ \textbf{Entangled Attention}: Existing works feed both the visual and textual features into the same set of stacked transformers. In this manner, the same set of attention matrices are used to transform both visual and textual embeddings. 
    We denote such type of attention mechanism as \textit{Entangled Attention}, because the visual and textual embeddings are projected to \textit{one common latent space}. 
    Despite the pre-training on image-sentence pairs, the initial parameters of current systems \cite{Zhou2019VLP,Su2019vlbert,Li2019VisualBERT,Li2019unicoder-VL,chen2019UNITER} are usually directly inherited from BERT \cite{devlin2018bert} or BERT-based models, e.g., UniLM \cite{Dong2019unilm}, which is optimized for language modeling only\footnote{Some works, e.g., VL-BERT \cite{Su2019vlbert}, attempt to pre-train the V-L systems on a large amount of text-only datasets in the initial pre-training stage.}, to boost the performance. On one hand, it could be in-appropriate to apply the parameters that are trained in language modality to the features in visual modality; On the other hand, the ability of language modeling brought by the original BERT could be somewhat affected by the introduced visual modality. 
    Besides, during parameter optimization, the model needs to consider the intra-relationships of both visual and textual embeddings, as well as the inter-relationships across V-L embeddings. Modeling these three kinds of relations with only one shared set of attention matrices could be insufficient.
    
    $\bullet$ \textbf{Modality Gap}: Most V-L systems only use the region-of-interests (RoIs) / video frames as the visual features. Although the transformers are proved to be effective in mining correlations, there are still huge gaps between the visual and language modalities.
    
    $\bullet$ \textbf{Generation Task}: Most systems can only be fine-tuned directly on classification tasks, lacking the capability to handle generation tasks. Although VideoBERT \cite{sun2019VideoBERT} and CBT \cite{Sun2019CBT} have been proposed to support the generation tasks in video domain, they have to train a separate video-to-text decoder to perform the generation tasks, because they pre-train the V-L systems only as encoders.

To tackle the above three concerns, in this paper, we present the DiMBERT, which takes both visual features (i.e., RoIs and visual concepts) from images and textual features (i.e., sentence words) from sentences as input, and then applies a single cross-modal Transformer to learn vision-language grounded representations. In particular, we propose Disentangled Multimodal-Attention (DiM) module to explicitly disentangle visual and textual modalities. In implementations, DiM module introduces separate projection matrices to project visual and textual modalities into their corresponding visual and textual latent spaces. Following common practice \cite{Alberti2019B2T2,Zhou2019VLP,Su2019vlbert,Li2019VisualBERT,Li2019unicoder-VL,chen2019UNITER}, the weights of textual projection matrices are initialized with the pre-trained parameters from BERT \cite{devlin2018bert}, while the weights of visual projection matrices are trained from scratch.

To enhance the correlation between visual and textual modalities, we introduce the visual concepts \cite{fang2015captions}, which capture a wide range of high-level visual semantic information from images \cite{yao2017boosting,Pan2017Transferred,Li2019VSCMR}. The visual concepts transform visual features to a set of words describing \textit{object} (e.g., \textit{cat}), \textit{attribute} (e.g., \textit{small}) and \textit{relationship} (e.g., \textit{standing}) of images, which 1) provide a more semantic representation of visual information and thus help shorten the gap between vision and language modalities; 2) contain rich visual semantics and thus help understand vision and language effectively.

Inspired by the work of \cite{Zhou2019VLP}, we pre-train the proposed DiMBERT on the Conceptual Captions \cite{Sharma2018conceptual} with two unsupervised language modeling objectives: bidirectional language modeling \cite{devlin2018bert} and sequence-to-sequence language modeling \cite{Dong2019unilm}, where the latter enables the direct fine-tuning of DiMBERT on generation tasks.
We conduct comprehensive experiments and systematic analysis on three tasks: image captioning \cite{chen2015microsoft}, visual storytelling \cite{visualstorytelling} and referring expressions \cite{Kazemzadeh2014GRE}. The proposed DiMBERT sets new state-of-the-arts on three tasks (over four benchmark datasets) and the DiM module boosts the performance of various pre-trained V-L models on referring expressions task, which validate our motivation and corroborate the effectiveness and universality of our approach.

\section{Related Work}

Our work relates to the vision-and-language problems, the joint representations of vision and language (a.k.a. V-L representations) and the efforts in developing pre-trained models.

\subsection{Vision-and-Language Problems}
Vision-and-language (V-L) problems, which including image captioning \cite{chen2015microsoft}, visual storytelling \cite{visualstorytelling}, referring expression \cite{Kazemzadeh2014GRE} and image caption retrieval \cite{Nam2017icretrieval}, and others, have drawn remarkable attention in both natural language processing and computer vision. These tasks combine image and language understanding together at the same time, are tough yet practical. 
However, current works usually design a task-specific model to deal with one single task at one time. In this paper, we propose a generic framework to conduct various V-L tasks and further improve the performance of each task.

\subsection{V-L Representations}
For a variety of vision-and-language problems, an important goal is to understand the image and language despite their different application scenarios, which justifies the acquisition of fine-grained V-L representations. In the literature, to represent the images, visual features extracted by CNNs or Region-CNNs are most-widely used \cite{xu2015show,anderson2018bottom}, while visual concepts consisting of a set of textual words are also proposed \cite{fang2015captions}. 
To represent the language, textual features extracted by RNNs or off-the-shelf NLP models are most-widely used \cite{elman1990finding,hochreiter1997long,devlin2018bert}.
Therefore, in previous task-specific V-L models, the features derived from off-the-shelf computer vision and NLP models are combined in an ad-hoc way to acquire the V-L representations for specific tasks. Model training is performed on the dataset for the specific task only, without any generic V-L pre-training. 
As a result, these task-specific models, which are directly trained for the specific target task, may well suffer from overfitting when the data for the target task is scarce.

\subsection{Pre-trained Models}

In computer vision (CV), pre-trained models, such as ResNet \cite{he2016deep}, VGG \cite{simonyan2014very} and GoogLeNet \cite{Szegedy2015Going}, pre-trained on ImageNet \cite{Deng2009ImageNet}, have achieved early successes in promoting various downstream CV tasks. Transformer-based pre-trained NLP models, such as BERT \cite{devlin2018bert}, XLNet \cite{Yang2019XLNet} and RoBERTa \cite{Dong2019unilm}, have also achieved great success in advancing the state-of-the-arts for a wide range of NLP tasks. Recently, UniLM \cite{Dong2019unilm}, which is adapted from the BERT architecture, has been proposed to enable BERT to work for both natural language understanding and generation tasks.

Most recently, several pre-trained V-L models \cite{sun2019VideoBERT,Sun2019CBT,lu2019vilbert,Tan2019LXMERT,Alberti2019B2T2,Zhou2019VLP,Su2019vlbert,Li2019VisualBERT,Li2019unicoder-VL,chen2019UNITER,Li2020Oscar,Xia2020XGPT,Miyazawa2020lamBERT,Gan2020Adversarial,Huang2020Pixel-BERT,Kim2021ViLT,Cho2020X-LXMERT,Li2020HERO,Ging2020COOT,Wang2020MiniVLM,Zhang2021VinVL,Lei2021ClipBERT,Li2020Closer,Jia2021Scaling,Li2020Weakly,Yu2020ERNIE-ViL,Chen2021VisualGPT,Li2020UNIMO,Huang2020M3P,Caglayan2021Cross,Bugliarello2020Multimodal,Xing2021KM-BART,Luo2020UniViLM,Cao2020Behind,Lu201912_in_1,Feng2020CodeBERT,Zhu2020ActBERT,Gao2020FashionBERT,2020ConceptBert,Khan2020MMFT-BERT,Burns2020Learning,Murahari2020Large} have been proposed to learn vision-language representations for various V-L tasks (Table~\ref{tab:compare} summarizes some representative works). However, most existing works do not consider to learn such representations by explicitly disentangling multi-modalities and incorporating visual concepts, and are unable to perform downstream generation tasks directly.
It is worth noticing that, VideoBERT \cite{sun2019VideoBERT}, CBT \cite{Sun2019CBT} and VLP \cite{Zhou2019VLP} proposed recently, are capable of performing generation tasks, and are thus the most relevant works to our approach. 
However, they were still entangling visual and textual modalities, and did not attempt to take the visual concepts into consideration.
Besides, for VideoBERT \cite{sun2019VideoBERT} and CBT \cite{Sun2019CBT}, they pre-train the systems only as encoders to learn V-L representations, so they have to train a separate decoder for the generation tasks.
For VLP \cite{Zhou2019VLP}, our DiMBERT outperforms it on image captioning tasks and our superiority is further validated on a long text generation task, i.e., visual storytelling \cite{visualstorytelling}.

\section{Approach}

In this section, we first introduce our model in detail. Next, we describe the pre-training tasks to learn vision-language grounded representations.
Figure~\ref{fig:overview} gives an overview of our DiMBERT.

\subsection{Model Overview}
\begin{figure*}[t]
\includegraphics[width=1\linewidth]{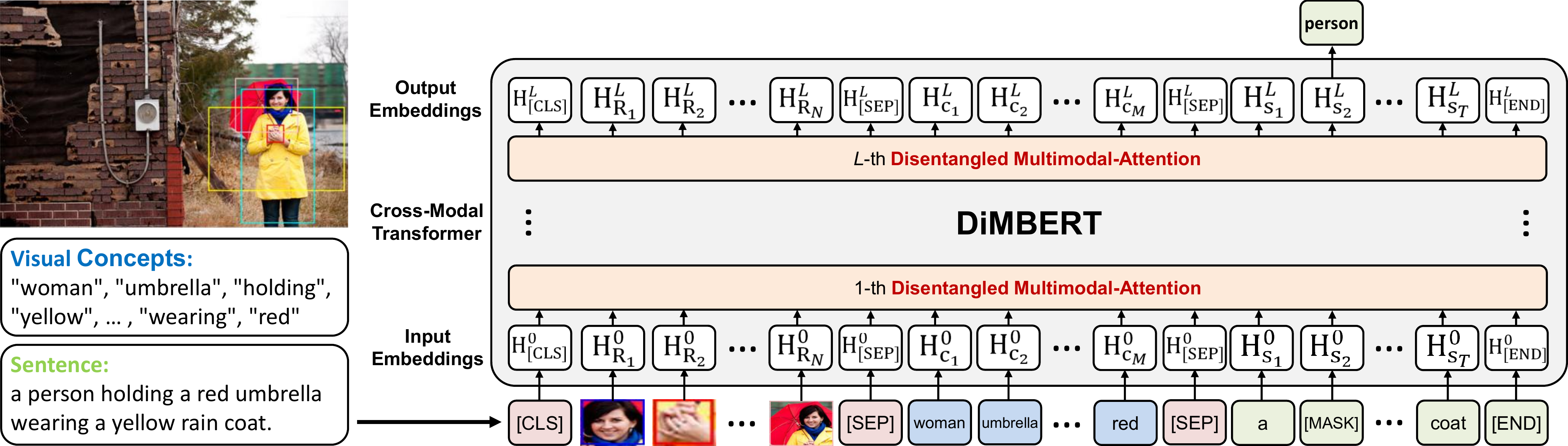}
\caption{Illustration of the proposed DiMBERT, which consists of a single cross-modal Transformer \cite{Su2019vlbert,ashish2017attention} and takes visual features (i.e., RoIs and visual concepts), textual features (i.e., sentence words) and four special tokens (i.e., [CLS], [SEP], [MASK] and [END]) as input. In particular, we introduce the Disentangled Multimodal-Attention to implement the Transformer.}
\label{fig:overview}
\end{figure*}

As shown in Figure~\ref{fig:overview}, our DiMBERT consists of three parts: 1) the embeddings of input visual features from images and textual features from sentences; 2) a single cross-modal transformer to learn the alignments and relationships between visual and textual modalities; and 3) the embeddings of learned V-L representations.

\subsubsection{Input Embeddings}

There are 4 types of input embeddings: RoIs, visual concepts, sentence words and four special tokens.

\smallskip\noindent\textbf{RoI Embedding.}
In our approach, we use $N\,$=$\,36$ RoIs for each input image. RoIs are extracted by a variant of Faster R-CNN \cite{ren2015faster} with ResNeXt-101 FPN backbone \cite{Xie2017ResNeXt}, which is pre-trained on Visual Genome \cite{krishna2017visualgenome}, following \cite{anderson2018bottom}.

Specifically, the appearance feature $\vec{r}_i \in \R^{d_r}$ is the extracted region feature, which is the output of \textit{fc6} layer.
The visual geometry feature $\vec{g}_i \in \R^{d_g}$ is used to encode the geometry location of the RoI in the input image, where 
$\footnotesize \vec{g}_i = \left(\frac{x_{\mathrm{TL}}}{W}, \frac{y_{\mathrm{TL}}}{H}, \frac{x_{\mathrm{BR}}}{W}, \frac{y_{\mathrm{BR}}}{H}, A_r \right)$, in which $\left(x_{\mathrm{TL}}, y_{\mathrm{TL}}\right)$ and $\left(x_{\mathrm{BR}}, y_{\mathrm{BR}}\right)$ 
denote the coordinates of the top-left and bottom-right corner, respectively, of the region bounding box; $W$, $H$ are the width and height of the input image; and $A_r$ represents the relative area, i.e., the area ratio of RoI bounding box to the entire image. 
Besides, following \cite{lu2018neural,Zhou2019GVD}, to enrich region features, we inject the region class information $\vec{c}_{i}$ into $\vec{g}_i$, defined as:
$g^*_{i} = \left[\text{LN}\left(W_{g} g_{i}\right); \text{LN}\left(W_{c} c_{i}\right)\right]$,
where [;] and LN stand for concatenation and layer normalization \cite{ba2016layernormalization}, respectively; $W_{g}$ and $W_{c}$ are learnable parameters; $\vec{c}_i \in \R^{d_c}$ is the prediction scores (probabilities) of region object label, where $d_c\,$=$\,1600$ is the number of object categories.
The segment embedding is used to indicate which input segment it belongs to. For RoIs, we adopt the embedding of [RoI] token $\E_{\text{[RoI]}}$ as segment embedding.
Finally, the RoI embedding $H_{{R}_i} \in \R^{d_{\text{model}}}$ is calculated by a weighted sum of these aforementioned embeddings.

\smallskip\noindent\textbf{Visual Concept Embedding.}
Visual concepts contain rich visual semantics, and have been used to provide explicit high-level semantic information of an image \cite{wu2016what}. Following \cite{fang2015captions}, we adopt a weakly-supervised approach of Multiple Instance Learning \cite{zhang2006multiple} to build the visual concepts extractor, which is trained on the MSCOCO caption dataset for 1,000 visual concepts.
For each image, only the top $M\,$=$\,20$ visual concepts are selected.
We sort these extracted visual concepts by prediction scores, which means that the position embedding of each visual concept indicates its relevance to the image. Following BERT, the visual concepts use the Word-Piece embeddings \cite{Wu2016Google}.
They will then be further combined with position embeddings and a segment token [CEP]. The result is denoted as the visual concept embedding $H_{{c}_i} \in \R^{d_{\text{model}}}$.

\smallskip\noindent\textbf{Sentence Word Embedding.}
Similar as in the visual concept embeddings, we tokenize the sentence into WordPieces \cite{Wu2016Google}. After that, a position embedding and a segment token [SEN] embedding are assigned to each sentence word, where the position embedding indicates its order in the input sentence. We denote each sentence word embedding as $H_{{s}_i} \in \R^{d_{\text{model}}}$.

At last, following BERT \cite{devlin2018bert}, we define a small set of special tokens: [CLS], [END], [SEP] and [MASK]. [CLS] and [END] are inserted at the first and last position, representing the start and the end of the sentence, respectively. [SEP] token is added as the boundary between two different input segments, and [MASK] token indicates the random masked-out word, which need to be predicted by DiMBERT based on all the other available elements, including RoIs, visual concepts and available sentence words.
Thus, the input embeddings can be written as:
\begin{align}
\footnotesize
H^{0}=\left\{H_{\text{[CLS]}}, H_{{R}}, H_{\text{[SEP]}}, H_{{c}}, H_{\text{[SEP]}}, H_{{s}}, H_{\text{[END]}}\right\} , 
\end{align}
where the $H_{{R}}$, $H_{{c}}$ and $H_{{s}}$ are the sets of related vectors.

\subsubsection{Single Cross-Modal Transformer}

Our DiMBERT is adapted from $\text{BERT}_\text{BASE}$, thus the backbone of our approach is the 12-layer ($L$) Transformer \cite{ashish2017attention}, with 768 hidden units ($d_\text{model}$) in each layer. In implementation, we propose the Disentangled Multimodal-Attention (DiM) to replace the Self-Attention (SA) in Transformer. So we hereby describe the difference between DiM and SA.
First, we denote the intermediate representations of $l$-th layer as $H^{l}=\left\{H_{{R}}^{l}, H_{{c}}^{l}, H_{{s}}^{l}\right\}$ (the representations of four special tokens are omitted for conciseness). And we find that $H_{{R}}^{l}$ extracted from the image belongs to the visual modality $H_{\text{V}}^{l}$, while $H_{{c}}^{l}$ extracted from the image and $H_{{s}}^{l}$ extracted from the sentence belong to the textual modality $H_{\text{T}}^{l}$, which we write as $H^{l}=\left\{H_{\text{V}}^{l}, H_{\text{T}}^{l}\right\}$.

\smallskip\noindent\textbf{Self-Attention (SA).}
In current pre-trained V-L systems \cite{Su2019vlbert,Li2019VisualBERT,Li2019unicoder-VL,chen2019UNITER,Zhou2019VLP}, to learn the alignments and relationships between the visual modality $H_{\text{V}}^{l}$ and textual modality $H_{\text{T}}^{l}$, they adopt SA, which consists of $n\,$=$\,12$ parallel heads with each head $\text{SA}_i$ defined as:
\begin{align}
\footnotesize
\label{eq:SA}
\text{SA}_i(H_{\text{V}}, H_{\text{T}}) = \text{softmax}\left(\left[\begin{array}{c}H_{\text{V}} {W}^{q}_{\text{T}}\\ H_{\text{T}} {W}^{q}_{\text{T}} \end{array}\right]\left[\begin{array}{c}H_{\text{V}} {W}^{k}_{\text{T}}\\ H_{\text{T}} {W}^{k}_{\text{T}}\end{array}\right]^{T}\right)\left[\begin{array}{c}H_{\text{V}} {W}^{v}_{\text{T}} \\ H_{\text{T}} {W}^{v}_{\text{T}}\end{array}\right] , 
\end{align}
where ${W}^{q}_{\text{T}}, {W}^{k}_{\text{T}}, {W}^{v}_{\text{T}}$ are parameters initialized with pre-trained parameters from BERT, which are pre-trained on text data only. Besides, the divisor $\sqrt{{d}_{k}}$ (${d}_{k} = {d}_\text{model} / n$ = 64) is omitted in equations for conciseness, please see \cite{ashish2017attention,devlin2018bert} for details. 
As we can see, SA first projects visual and textual modalities into \textit{one common latent space}, resulting in the entanglements between the two modalities.
After SA, they use a position-wise feed-forward network (FFN) \cite{ashish2017attention,devlin2018bert}, which keeps on processing the features with mixed modalities and thus loses the capability to learn the relationships between the two modalities \cite{Zhao2019MUSE}.
After that, the intermediate representations of ($l$+1)-th layer $H^{l+1}$ are obtained by: 
$H_{\text{V}}^{l+1}, H_{\text{T}}^{l+1}=\text{FFN}(\text{SA}(H_{\text{V}}^{l}, H_{\text{T}}^{l}))$.

\smallskip\noindent\textbf{Disentangled Multimodal-Attention (DiM).}
As we can see, due to the entanglements between the visual and textual modalities in SA, the models have to devote most of its capability on disentangling them, which makes it hard for systems to learn the relationships between visual and textual modalities efficiently.
To this end, we propose Disentangled Multimodal-Attention (DiM) to explicitly disentangle visual and textual modalities. 
In implementation, the DiM also consists of $n$ parallel heads but with each head $\text{DiM}_i$ defined as:
\begin{align}
\footnotesize
\label{eq:DiM}
\text{DiM}_i(H_{\text{V}}, H_{\text{T}}) = \text{softmax}\left(\left[\begin{array}{c}H_{\text{V}} {W}^{q}_{\text{V}}\\ H_{\text{T}} {W}^{q}_{\text{T}} \end{array}\right]\left[\begin{array}{c} H_{\text{V}} {W}^{k}_{\text{V}}\\ H_{\text{T}} {W}^{k}_{\text{T}}\end{array}\right]^{T}\right)\left[\begin{array}{c}H_{\text{V}} {W}^{v}_{\text{V}} \\ H_{\text{T}} {W}^{v}_{\text{T}}\end{array}\right]
\end{align}
where ${W}^{q}_{\text{T}}, {W}^{k}_{\text{T}}, {W}^{v}_{\text{T}}$ are initialized with UniLM \cite{Dong2019unilm} parameters pre-trained on text data only; and the ${W}^{q}_{\text{V}}, {W}^{k}_{\text{V}}, {W}^{v}_{\text{V}}$ are new learnable parameters (randomly initialized).
After that, following SA, we obtain the intermediate representations of ($l$+1)-th layer $H^{l+1}$ by: 
$H_{\text{V}}^{l+1}, H_{\text{T}}^{l+1}=\text{FFN}(\text{DiM}(H_{\text{V}}^{l}, H_{\text{T}}^{l}))$.

The reason that we adopt the proposed Disentangled Multimodal-Attention is to learn the alignments and relationships between visual and textual modalities in a more efficient way.

\subsubsection{Output Embeddings}

After multiple DiM layers, we use the output of last layer as the output embeddings. ($H^L_R$, $H^L_c$ and $H^L_s$ denotes RoIs, concepts and sentence representations, respectively)

\subsection{Pre-Training Tasks}
 
In our work, we pre-train DiMBERT on the training split \cite{Zhou2019VLP} of a large scale image-sentence dataset: i.e., Conceptual Captions dataset \cite{Sharma2018conceptual}, which contains around 3.3M image-sentence pairs. To pre-train DiMBERT, we introduce two unsupervised language modeling tasks, which are adapted from the masked language modeling (MLM) task: 1) bidirectional language modeling (BLM) \cite{devlin2018bert}: which learns to predict the randomly masked sentence words based on all available input information, i.e., the visual and textual features; 2) sequence-to-sequence language modeling (S2SLM) \cite{Dong2019unilm}: which learns to predict the randomly masked sentence words based on partial input information, i.e., all the visual features and the sentence words on the left side of the word to be predicted in the sentence, which satisfies the auto-regressive property and enables our DiMBERT to perform downstream generation tasks.
In this section, we will describe these pre-training task in detail.

\smallskip\noindent\textbf{Masked Language Modeling (MLM).}
Following BERT, during the pre-training stage, we randomly mask out the input sentence words with 15\% probability, replacing the word with 80\%, 10\% and 10\% probabilities of [MASK] token, random word and original word, respectively.
Thus, the objective of MLM is to predict the randomly masked sentence word based on the available information. 
We denote trainable parameters as $\theta$, and a pair $({w}_{\text{m}},{e}_{\text{a}})$ of input with the masked word as ${w}_{\text{m}}$, and all available elements as ${e}_{\text{a}}$, which are sampled from the training set $D$.
The MLM is trained via minimizing negative log likelihood, defined as:
\begin{align}
\mathcal{L}_{\mathrm{MLM}}(\theta)=-E_{({w}_{\text{m}}, {e}_{\text{a}}) \sim D} \log \left(p_{\theta}\left({w}_{\text{m}} | {e}_{\text{a}}\right)\right) ,
\end{align}
where the masked tokens are predicted as a classification problem.

\smallskip\noindent\textbf{Bidirectional Language Modeling (BLM) and Seq-to-Seq Language Modeling (S2SLM).}
The main difference between two LM tasks is the different portion of available information that can be used to predict the masked word ${w}_{\text{m}}$.
For BLM, as in BERT \cite{devlin2018bert}, the model is allowed to use all the input embeddings on both left and right side of the [MASK] token. Thus ${e}_{\text{a}}$ consists of all RoIs $\vec{R}$, all visual concepts $\vec{c}$ and all other sentence words $\vec{s}_{\backslash {w}_{\text{m}}}$. The loss function is defined as:
\begin{align}
\mathcal{L}_{\mathrm{BLM}}(\theta)=-E_{({w}_{\text{m}}, {e}_{\text{a}}) \sim D} \log \left(p_{\theta}\left({w}_{\text{m}} | \vec{R}, \vec{c}, \vec{s}_{\backslash {w}_{\text{m}}}\right)\right) .
\end{align}
For S2SLM, in order to enable the encoder-decoder and auto-regressive properties in self-attention layer, each visual elements (i.e., RoIs and visual concepts) in the first two segments is only allowed to attend to other visual elements within the first two segments, which constitutes a visual encoder; for predicting the masked sentence word, only the left side elements of the [MASK] token can be used, which constitutes a sentence decoder. If we denote the position of [MASK] token in the input sentence as $t$, then the usable elements are all the RoIs $\vec{R}$, all the visual concepts $\vec{c}$ and all the left side of the [MASK] token in the sentence $\vec{s}_{1:t}$, and the loss function is:
\begin{align}
\mathcal{L}_{\mathrm{S2SLM}}(\theta)=-E_{({w}_{\text{m}}, {e}_{\text{a}}) \sim D} \log \left(p_{\theta}\left({w}_{\text{m}} | \vec{R}, \vec{c}, \vec{s}_{1:t}\right)\right) .
\end{align}

In implementation, the two LM tasks, which are alternated with random sampling, participate in the pre-training stage at a ratio of 25\% and 75\%, respectively.
We pre-train DiMBERT on 8 GPUs (Tesla V100) with a batch size of 512 for 30 epochs. We use the Adam optimizer \cite{kingma2014adam} with initial learning rates of 3e-4. 
Through pre-training on Conceptual Captions dataset, our model is capable to learn vision-language grounded representations. 
Following VisualBERT \cite{Li2019VisualBERT}, to let our DiMBERT better adapt to the downstream target domains, we further pre-train DiMBERT using the data from downstream tasks. Eventually, we get the final pre-trained model by averaging the last 20 checkpoints.

\begin{figure*}[t]
\includegraphics[width=1\linewidth]{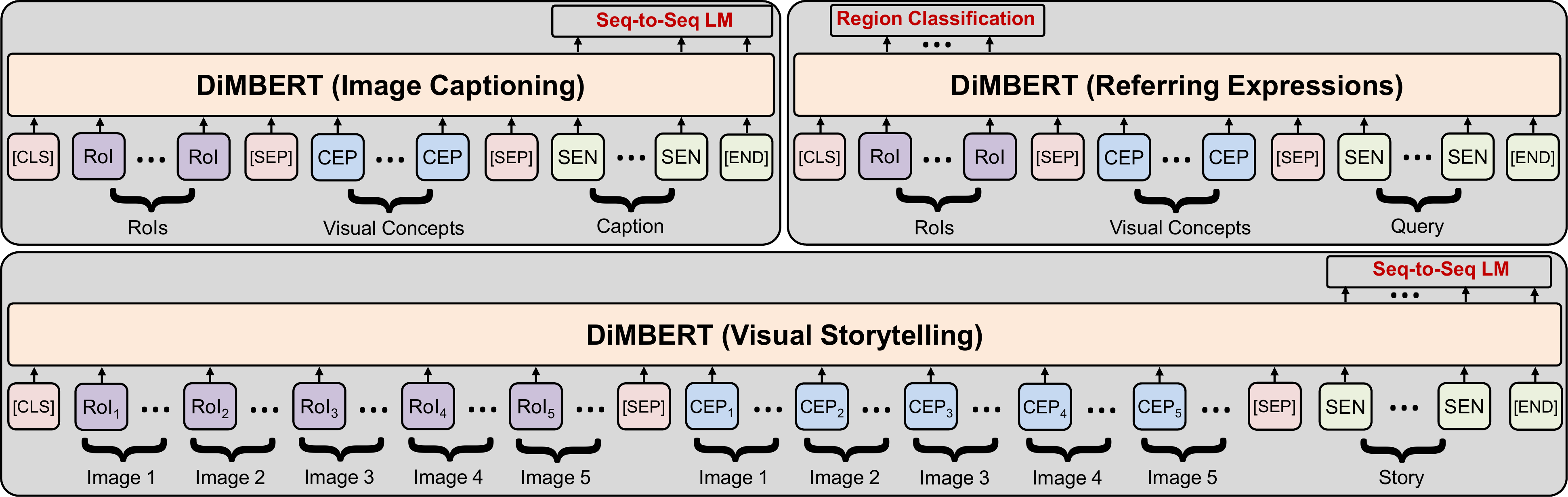}
\caption{Illustration of fine-tuning the pre-trained DiMBERT on various vision-and-language downstream tasks, including two generation tasks, i.e., image captioning and visual storytelling, and a classification task, i.e., referring expressions.}

\label{fig:fine-tuning}
\end{figure*}

\section{Experiments}
\label{sec:exp}

We evaluate DiMBERT on three  representative vision-and-language tasks, i.e., two generation tasks (image captioning \cite{chen2015microsoft} and visual storytelling \cite{visualstorytelling}) and a classification task (referring expressions \cite{Kazemzadeh2014GRE}).

\subsection{Image Captioning}
The task of image captioning aims to generate a descriptive sentence for an input image and has received extensive research interests.

\smallskip\noindent\textbf{Datasets and Metrics.}
We use the popular Flickr30k \cite{young2014image} and MSCOCO \cite{chen2015microsoft} datasets to evaluate our reported results. The datasets contain 31,783 images and 123,287 images, respectively, with 5 sentences paired to each image. To make fair comparisons \cite{Huang2019AoA,Yang2019Auto-Encoding,Zhou2019VLP}, we use the widely-used splits in the work of Karpathy and Li \cite{karpathy2014deep} to report our results.
As a result, there are 5,000 images each in the validation set and the test set for MSCOCO, and 1,000 images as for Flickr30k.

We test the model performance with MSCOCO captioning evaluation toolkit \cite{chen2015microsoft}, which reports the widely-used automatic evaluation metrics SPICE \cite{anderson2016spice}, CIDEr \cite{vedantam2015cider}, ROUGE \cite{lin2004rouge}, METEOR \cite{lin2003automatic,banerjee2005meteor} and BLEU \cite{papineni2002bleu}.
SPICE is based on scene graph matching and CIDEr is based on n-gram matching. They both incorporate the consensus of a reference set for an example. These two metrics are specifically designed for the evaluation of image captioning systems. ROUGE is proposed for automatic evaluation of the extracted text summarization. METEOR and BLEU are originally designed for machine translation evaluation.

\smallskip\noindent\textbf{Fine-Tuning and Inference.}
Figure~\ref{fig:fine-tuning} illustrates the details of fine-tuning. As we can see, we apply the Seq-to-Seq Language Modeling task to fine-tune (cross-entropy optimization) the pre-trained DiMBERT on the image captioning task. The pre-trained DiMBERT is fine-tuned on 8 GPUs with a batch size of 512 for 30 epochs. We use the learning rate of 3e-5 and 1e-4 for parameter optimization on Flickr30k and MSCOCO datasets, respectively. 
Furthermore, for fair comparisons with state-of-art works \cite{Huang2019AoA,Yang2019Auto-Encoding} on the MSCOCO dataset, we further perform CIDEr-based training objective using reinforcement training \cite{rennie2017self} with a learning rate of 1e-6. 

In the inference stage, we initialize the input of model with \{[CLS], RoIs, [SEP], Concepts, [SEP], [MASK]\}, then the model will generate a \textit{word}$_1$ from the position of [MASK] token. Next, DiMBERT takes the \{[CLS], RoIs, [SEP], Concepts, [SEP], \textit{word}$_1$, [MASK]\} as input. The entire inference process repeats such generation until DiMBERT outputs an [END] token.
Following common practice \cite{yao2018exploring,Yang2019Auto-Encoding,liu2021PPKED,Huang2019AoA,Zhou2019VLP,Yang2021NACF},
we apply beam search with beam size = 3 during inference.

\smallskip\noindent\textbf{Results.}
We compare our DiMBERT with two types of existing works: 1) The state-of-the-art task-specific models like GVD \cite{Zhou2019GVD} on Flickr30k and AoANet \cite{Huang2019AoA} on MSCOCO image captioning datasets. 2) The pre-training based models, like VLP \cite{Zhou2019VLP}. The results are shown in Table~\ref{tab:result}. As we can see, the proposed DiMBERT outperforms all baselines across all metrics over the board, which demonstrates the capability of DiMBERT to achieve consistent performance gains over different datasets.

We also evaluate our DiMBERT on the online MSCOCO evaluation server\footnote{\url{https://competitions.codalab.org/competitions/3221\#results}}, where the ground truth captions are not available. We compare with the top-performing entries on the leaderboard whose methods are published, which including AoANet \cite{Huang2019AoA}, SGAE \cite{Yang2019Auto-Encoding}, ETA \cite{Li2019ETA}. For online evaluation, nearly all of the recent submitted systems use model ensemble \cite{Huang2019AoA,jiang2018recurrent,anderson2018bottom,Yang2019Auto-Encoding,yao2018exploring}. From Table~\ref{tab:res-server}, we can find that our DiMBERT is able to outperform these state-of-the-art models across all metrics over the board, in a single model submission. It further demonstrates the effectiveness of the proposed DiMBERT.

\begin{table*}[t]

\footnotesize
\setlength{\tabcolsep}{1pt}

\caption{Comparisons with the state-of-the-art task-specific models and pre-training models on various downstream tasks, i.e., image captioning (Flickr30k and MSCOCO datasets (with CIDEr optimization)), visual storytelling (VIST dataset) and referring expression (RefCOCO+ dataset). B-4, M, R, C
and S are short for BLEU-4, METEOR, ROUGE-L, CIDEr and SPICE, respectively. For referring expression task,  following common practice \cite{Su2019vlbert,chen2019UNITER}, we evaluate on both ground-truth RoIs (val, testA, testB) and detected boxes ($\text{val}^{d}$, $\text{testA}^{d}$, $\text{testB}^{d}$) provided by \cite{Yu2018MAttNet}. The models of referring expression are evaluated in terms of accuracy (\%).} 
\label{tab:result}

\begin{tabular}{@{}l l c c c c c c c c c c c c c c c c c c c c@{}}
\toprule
& \multirow{2}{*}[-3pt]{Methods}  & \multicolumn{6}{c}{RefCOCO+} & \multicolumn{4}{c}{VIST} &  \multicolumn{5}{c}{Flickr30k Image Captioning} &  \multicolumn{5}{c}{MSCOCO Image Captioning} \\ \cmidrule(lr){3-8} \cmidrule(lr){9-12} \cmidrule(lr){13-17} \cmidrule(l){18-22} & & val & testA & testB & $\text{val}^{d}$ & $\text{testA}^{d}$ & $\text{testB}^{d}$ & B-4  & M   & R  & C   & B-4  & M   & R  & C & S  & B-4  & M   & R  & C & S   \\ \midrule [\heavyrulewidth]

\multirow{6}{*}{\rotatebox{90}{\begin{tabular}[c]{@{}c@{}} Task-Specific \\ Models \end{tabular}}} 
& MAttNet \cite{Yu2018MAttNet} & 71.0 & 75.1 & 66.1 & 65.3  & 71.6  &56.0 & - & - & - & - & - & - & - & - & - & -  & - & - & -  & - \\
& AREL \cite{Wang2018AREL} & - & - & -& - & - & - & 14.1 & 35.0 & 29.5 & 9.4 & - & - & - & - & - & - & - & - & - & - \\
& VSCMR \cite{Li2019VSCMR}  & - & - & -& - & - & - &14.3& 35.5 &30.2 &9.0 & - & - & - & - & - & - & - & - & - & -  \\
& GVD \cite{Zhou2019GVD} & - & - & - & - & - & - & -& - & - & -  & 27.3 & 22.5 & -  & 62.3 &16.5 & - & - & - & - & - \\
& SGAE \cite{Yang2019Auto-Encoding}  & -& -& -& -& -& -& -& - & - & - & - & - & - & - & - & 39.0 & 28.4 & 58.9 & 129.1 & 22.2 \\
& AoANet \cite{Huang2019AoA} & -& -& -& -& -& - & - & -& -& - & - & - & - & - & - & 38.9 & 29.2 & 58.8 & 129.8 & 22.4 \\

\midrule [\heavyrulewidth]
\multirow{7}{*}{\rotatebox{90}{\begin{tabular}[c]{@{}c@{}} Pre-Training \\ Models \end{tabular}}} 
& ViLBERT \cite{lu2019vilbert} & - & - & - & 72.3 & 78.5 & 62.6  & - & - & -  & - & - & - & - & -  & - & - & - & - & -  & -  \\
& $\text{VL-BERT}_\text{LARGE}$ \cite{Alberti2019B2T2} & 80.3 & 83.6 & 75.5 & 72.6  & 78.6 & 62.3  & - & - & -  & - & - & - & - & -  & - & - & - & - & -  & - \\
& $\text{UNITER}_\text{LARGE}$  \cite{chen2019UNITER} & 84.0 & 85.9 & 78.9 & 74.9 & 81.4 & 65.4  & - & - & -  & - & - & - & - & -  & - & - & - & - & -  & - \\
& XGPT \cite{Xia2020XGPT} & -  & -  & -  & - & - & - & - & -  & -  & - & 31.8 & 23.6 & - & 70.9 & 17.6 & 37.2 & 28.6 & - & 120.1 & 21.8  \\
& VLP  \cite{Zhou2019VLP} & -  & -  & -  & - & - & - & - & -  & -  & -  & 31.1 &23.0 & - &68.5 & 17.2 &  39.5 & 29.3 & - & 129.3 & 23.2  \\
\cmidrule(){2-22}

& DiMBERT [Ours] & \bf 84.6 & \bf 86.0 & \bf 79.7 & \bf 75.6 & \bf 81.6 & \bf 66.9 & \bf 15.3 & \bf 36.0 & \bf 31.3  & \bf 13.8 & \bf 32.4 & \bf 24.1  & \bf 50.7   & \bf 72.3  & \bf 17.9 & \bf 40.7 & \bf 29.7 & \bf 59.6 & \bf 135.3 & \bf 23.7 \\
\bottomrule
\end{tabular}

\end{table*}
\begin{table*}[t]
\centering
\footnotesize
\caption{Leaderboard performance on the online MSCOCO image captioning evaluation server. c5 means comparing to 5 references and c40 means comparing to 40 references. As we can see, our DiMBERT outperforms all  state-of-the-art models across all metrics over the board, including AoANet \cite{Huang2019AoA} that uses 4-model ensemble, in a single model submission.
\label{tab:res-server}}
\begin{tabular}{@{}l c c c c c c c c c c c c c c c@{}}
\toprule
\multirow{2}{*}[-3pt]{Methods}  
& \multicolumn{2}{c}{BLEU-1}
&  \multicolumn{2}{c}{BLEU-2}
&  \multicolumn{2}{c}{BLEU-3}
&  \multicolumn{2}{c}{BLEU-4}
&  \multicolumn{2}{c}{METEOR}
& \multicolumn{2}{c}{ROUGE-L}
& \multicolumn{2}{c}{CIDEr}  \\
\cmidrule(lr){2-3} \cmidrule(lr){4-5} \cmidrule(lr){6-7} \cmidrule(lr){8-9} \cmidrule(lr){10-11} \cmidrule(lr){12-13} \cmidrule(lr){14-15}

& c5 & c40 & c5 & c40 & c5 & c40 & c5 & c40 & c5 & c40 & c5 & c40 & c5 & c40 \\
\midrule
SCST   \cite{rennie2017self}            &78.1           & 93.7     & 61.9     & 86.0     & 47.0 &75.9 &35.2 &64.5 &27.0 &35.5 &56.3 &70.7 &114.7 &116.7     \\
Up-Down \cite{anderson2018bottom}  & 80.2     & 95.2 &64.1     & 88.8     &  49.1 &79.4 &36.9 &68.5 &27.6 &36.7 &57.1 &72.4 &117.9 &120.5 \\
CAVP \cite{liu2018context}  & 80.1     &  94.9 &64.7     & 88.8     &  50.0 &79.7 &37.9 &69.0 &28.1 &37.0 &58.2 &73.1 &121.6 &123.8 \\
ETA \cite{Li2019ETA} &81.2 &95.0 &65.5 &89.0 &50.9 &80.4 &38.9 &70.2 &28.6 &38.0 &58.6 &73.9 &122.1 &124.4 \\
RFNet \cite{jiang2018recurrent}   & 80.4     & 95.0 & 64.9 & 89.3 &  50.1 &80.1 &38.0 &69.2 &28.2 &37.2 &58.2 &73.1 &122.9 &125.1 \\
GLIED \cite{liu2019GLIED} &80.1 &94.6 &64.7 &88.9 &50.2 &80.4 &38.5 &70.3 &28.6 &37.9 &58.3 &73.8 &123.3 &125.6 \\
GCN-LSTM \cite{yao2018exploring} & - & - &65.5 &89.3 &50.8 &80.3 &38.7 &69.7 &28.5 &37.6 &58.5 &73.4 &125.3 &126.5 \\
SGAE \cite{Yang2019Auto-Encoding} &81.0 &95.3 &65.6 &89.5 &50.7 &80.4 &38.5 &69.7 &28.2 &37.2 &58.6 &73.6 &123.8 &126.5 \\
AoANet \cite{Huang2019AoA} &81.0 &95.0 &65.8 &89.6 &51.4 &81.3 &39.4 &71.2 &29.1 &38.5 &58.9 &74.5 &126.9 &129.6 \\
\midrule
DiMBERT [Ours] & \bf 81.7	& \bf 96.1	& \bf 66.4	& \bf 91.0	& \bf 51.8	& \bf 83.0	& \bf 39.7	& \bf 73.1	& \bf 29.5	& \bf 39.2	& \bf 59.3	& \bf 75.0	& \bf 129.9	& \bf 133.1 \\
\bottomrule
\end{tabular}

\end{table*}

\subsection{Visual Storytelling}
Visual storytelling task belongs to long text generation tasks. The goal is to generate a reasonable and coherent paragraph-level story based on the image stream. 

\smallskip\noindent\textbf{Datasets and Metrics.}
In visual storytelling, our reported results are evaluated on the VIST dataset \cite{visualstorytelling}, which contains 210,819 unique images in 10,117 Flickr albums. Each sample contains one story, describing five selected images. We follow the standard split \cite{Wang2018AREL} for fair comparisons. There are 40,098 / 4,988 / 5,050 images for training / validation / testing, respectively. Following common practice \cite{Li2019VSCMR,Wang2018AREL,Yang2019Storyteller}, we adopt four evaluation metrics, including BLEU, ROUGE, METEOR and CIDEr, for the evaluation of our approach.

\smallskip\noindent\textbf{Fine-Tuning and Inference.}
For the visual storytelling task, we adopt the same fine-tuning and inference strategy as the image captioning task. These two tasks differ solely in the input format. 
As illustrated in Figure~\ref{fig:fine-tuning}, the visual storytelling task takes five images as input. For each image, we first extract corresponding RoIs and concepts. Then we feed the embeddings of these RoIs and concepts into DiMBERT from Image 1 to Image 5. 

\smallskip\noindent\textbf{Results.}
We choose two recently proposed state-of-the-art models AREL \cite{Wang2018AREL} and VSCMR \cite{Li2019VSCMR} for comparison. As shown in Table~\ref{tab:result}, the DiMBERT outperforms the AREL and VSCMR by a large margin of 46.8\% and 53.3\% in terms of CIDEr scores, respectively, and sets a new state-of-the-art, which shows that the proposed DiMBERT also works well on generating long texts.

\subsection{Referring Expressions}
Referring Expressions belongs to classification tasks which aims to locate a target image region given a textual query.

\smallskip\noindent\textbf{Datasets and Metrics.}
We evaluate our model on RefCOCO+ dataset \cite{Kazemzadeh2014GRE}, which consists of 141k expressions for 50k referred objects in 20k images in the MSCOCO dataset \cite{lin2014microsoft}.
The referring expressions in RefCOCO+ are forbidden from using absolute location words, e.g. right car. Therefore the referring expressions focus on purely appearance-based descriptions. 
RefCOCO+ is split into train, validation and two test sets (testA and testB). 
Specifically, images containing multiple people are in testA set, while images containing multiple objects of other categories are in testB set.
Following common practice \cite{Su2019vlbert,chen2019UNITER}, we evaluate on both ground-truth RoIs (val, testA, testB) and detected boxes ($\text{val}^{d}$, $\text{testA}^{d}$, $\text{testB}^{d}$) provided by \cite{Yu2018MAttNet}. We choose MAttNet \cite{Yu2018MAttNet}, ViLBERT \cite{lu2019vilbert}, VL-BERT \cite{Su2019vlbert} and UNITER \cite{chen2019UNITER} for comparison, where the first one MAttNet is the state-of-the-art task-specific model, while the rest are pre-training models. 
The models are evaluated in terms of accuracy (\%).

\smallskip\noindent\textbf{Fine-Tuning and Inference.}
As illustrated in Figure~\ref{fig:fine-tuning}, we add a linear layer on the $H^L_V$ to output the classification scores for all the input RoIs, and take the RoI with the highest score as the final prediction. 
DiMBERT is fine-tuned under a binary cross-entropy loss on 8 GPUs with a batch size of 256 for 20 epochs.

\smallskip\noindent\textbf{Results.}
Table~\ref{tab:result} shows that our DiMBERT outperforms all models, which validates the effectiveness of DiMBERT in referring expressions task.
In particular, DiMBERT outperforms $\text{VL-BERT}_\text{LARGE}$ and $\text{UNITER}_\text{LARGE}$ which are developed from a larger model $\text{BERT}_\text{LARGE}$ \cite{devlin2018bert}, while DiMBERT is adapted from $\text{BERT}_\text{BASE}$ \cite{devlin2018bert}. In other word, DiMBERT can achieve higher accuracy with much less parameters.
Furthermore, in addition to Conceptual Captions dataset~\cite{Sharma2018conceptual}, $\text{UNITER}_\text{LARGE}$ uses extra a large amount of image-sentence pairs (see Table~\ref{tab:compare}), e.g., VG Captions \cite{krishna2017visualgenome} and SBU Captions \cite{Ordonez2011Im2Text}, to pre-train the model. This proves the efficiency of DiMBERT in pre-training datasets and parameters.

In all, these results suggest that the DiMBERT can be applied to a wide range of downstream tasks, no matter what the type of task is (e.g., classification task and generation task). More encouragingly, our approach outperforms existing published state-of-the-art models, including the task-specific models and pre-training models, across all aforementioned tasks, which further confirms the effectiveness and universality of the proposed DiMBERT.

\begin{table*}[t]

\footnotesize
\setlength{\tabcolsep}{2.2pt}

\caption{Ablation study of our proposed DiMBERT which is performed on MSCOCO (with cross-entropy optimization only) and RefCOCO+.}
\label{tab:quantitative_analysis}
\begin{tabular}{@{}c c c c c c c c c c c c c c@{}}
\toprule
\multirow{2}{*}[-3pt]{Components}  &\multirow{2}{*}[-3pt]{\begin{tabular}[c]{@{}c@{}}  Init. from \\ BERT \cite{devlin2018bert} \end{tabular}} &\multirow{2}{*}[-3pt]{\begin{tabular}[c]{@{}c@{}}  Init. from \\ UniLM \cite{Dong2019unilm}  \end{tabular}} &\multirow{2}{*}[-3pt]{\begin{tabular}[c]{@{}c@{}} BLM \end{tabular}} &  \multirow{2}{*}[-3pt]{\begin{tabular}[c]{@{}c@{}} S2SLM \end{tabular}}   &  \multirow{2}{*}[-3pt]{\begin{tabular}[c]{@{}c@{}} Image-Sentence  \\ Relationship Prediction \end{tabular}}  &  \multirow{2}{*}[-3pt]{\begin{tabular}[c]{@{}c@{}} Masked Obj. \\ Prediction \end{tabular}}  &  \multirow{2}{*}[-3pt]{\begin{tabular}[c]{@{}c@{}} Masked ViCo \\ Prediction \end{tabular}}  & \multirow{2}{*}[-3pt]{\begin{tabular}[c]{@{}c@{}} DiM \end{tabular}} & \multicolumn{1}{c}{RefCOCO+} & \multicolumn{4}{c}{MSCOCO}     \\ 
\cmidrule(lr){10-10}  \cmidrule(l){11-14}  & & & & & &  & & & $\text{val}^{d}$ & B-4 & M & C & S   \\ \midrule [\heavyrulewidth]

Base  & & & & & & & & & 68.5 & 35.5 & 28.0 & 113.7 & 20.8 \\

\midrule

(1)  & $\surd$ & & & &  & & & &68.7 & 35.2 & 28.2 & 113.9 & 20.9  \\

(1)  & &$\surd$  & & & & & & &68.8 & 36.0 & 28.4 & 115.7 & 21.4  \\
\midrule

(2) &  &$\surd$ & $\surd$ & & &  & & &72.3 & 37.5 & 28.4 & 118.2 & 21.5\\

(2) & &$\surd$  & & $\surd$ & & &  & &71.5 & 37.8 & 28.5 & 118.8 & 21.7  \\

(2) & &$\surd$  & $\surd$ & $\surd$ & & & &  &  73.4 &38.0 & 28.5 & 119.3 & 21.7  \\
\midrule

(3)  & &$\surd$  & $\surd$ & $\surd$ &$\surd$ & &  & &72.6 & 36.3 & 28.3 & 116.4 & 21.6  \\

(3)  &  &$\surd$ & $\surd$ & $\surd$&  & $\surd$ & & &  73.3 & 35.8 & 28.2 &  114.9  & 21.0\\ 

(3)  &  &$\surd$ & $\surd$ & $\surd$& & &$\surd$ & & 71.5 & 36.4 & 28.4 &  115.6  & 21.3 \\ 

\midrule

DiMBERT & &$\surd$  & $\surd$ & $\surd$ & & & & $\surd$  &\bf 75.6 &\bf 38.1 &\bf 28.7 &\bf 123.4 &\bf 22.0  \\
\bottomrule
\end{tabular}

\end{table*}

\section{Analysis}
In this section, we conduct a series of analysis on a generation task, i.e., image captioning (MSCOCO), and a classification task, i.e., referring expressions (RefCOCO+), to provide some insights and answer the following questions: (1) What is the contribution of each component in DiMBERT? (2) What is the effect of Disentangled Multimodal-Attention module? (3) What is the contribution of visual concepts? (4) Why DiMBERT can adapt effectively to a wide range of downstream tasks? (5) Where does the actual improvement in the evaluation scores comes from? Please note that the performance on MSCOCO image captioning dataset reported in this section is different from the ones reported in Section~\ref{sec:exp}: As for fair comparisons with state-of-art works \cite{Huang2019AoA,Yang2019Auto-Encoding}, we further perform CIDEr-based training objective using reinforcement training \cite{rennie2017self}. While in this section, for simplicity, we directly do fine-tuning with cross entropy loss.

\subsection{Ablation Study}
In this section, we conduct the ablation analysis on RefCOCO+ and MSCOCO image captioning datasets. To analyze the effect of each component, we evaluate from the following three perspectives:

    $\bullet$ \textbf{Parameter Initialization}: the initial parameters inherited from BERT\cite{devlin2018bert} or UniLM~\cite{Dong2019unilm}.
    
    $\bullet$ \textbf{Language Model Pre-train Strategies}: the pre-train on the textual part of DiMBERT: Bidirectional Language Modeling (BLM) and Seq-to-Seq Language Modeling (S2SLM).
    
    $\bullet$ \textbf{Language-Vision Pre-train Strategies}: (1) Image-Sentence Relationship Prediction: predict whether an image and a sentence match each other, this task is introduced by LXMERT \cite{Tan2019LXMERT}; (2) Masked Object Prediction: predict the label of masked region; (3) Masked Visual Concept (ViCo) Prediction task, predict the masked visual concept.
    
Table~\ref{tab:quantitative_analysis} shows the results in which the base model denote the one trained from scratch on downstream tasks without parameter initialization and pre-train, DiMBERT denotes the full model proposed in this paper. We can find that:

$\bullet$ Initializing the parameters from existing models like BERT is beneficial, as the pre-trained language models could better capture the contextual representations and the structure of sentences.

$\bullet$ Comparing the results of (1) and (2) in Table~\ref{tab:quantitative_analysis}, both pre-training tasks Bidirectional Language Modeling and Seq-to-Seq Language Modeling in DiMBERT can facilitate referring expressions and image captioning. Pre-training only on the textual part of DiMBERT is helpful for downstream tasks.

$\bullet$ The introduction of Image-Sentence Relationship Prediction, Masked Object Prediction and Masked ViCo Prediction pre-training tasks actually hurts the performance. 
For Masked Object Prediction, it may due to the introduction of noise when there exists overlapped regions or wrongly detected labels \cite{Zhou2019VLP}. For Masked ViCo Prediction, it may due to masking and predicting wrongly extracted visual concepts, which mislead the model to learn relationships between visual and textual features.
For Image-Sentence Relationship Prediction, the unmatched image-sentence training pairs could hinder the training of other pre-training tasks \cite{Su2019vlbert,Zhou2019VLP}.

\begin{table}[t]

\footnotesize
\caption{Impact of DiM. ESABERT and DiMBERT represent the vanilla BERT model and ``BERT w/ DiM'', respectively.}
\label{tab:effect_DMA}
\begin{tabular}{@{}l c c c c c@{}}
\toprule
\multirow{2}{*}[-3pt]{Methods}  &RefCOCO+ & \multicolumn{4}{c}{MSCOCO}     \\ 
\cmidrule(lr){2-2}  \cmidrule(l){3-6}   & $\text{val}^{d}$ & B-4 & M & C & S  \\ \midrule [\heavyrulewidth]
ESABERT &73.4 & 38.0 & 28.5 & 119.3 & 21.7  \\ 
\ w/ DiM (DiMBERT)   &\bf 75.6  &\bf38.1 &\bf 28.7 &\bf 123.4 &\bf 22.0  \\

\bottomrule
\end{tabular}

\end{table}

\begin{table}[t]

\footnotesize
\caption{Analysis about the effectiveness and universality of DiM. We perform the analysis on the RefCOCO+ dataset.}
\label{tab:DiM}
\begin{tabular}{@{}l c c c|l c c c@{}}
\toprule
\multirow{2}{*}[-3pt]{ Methods } & \multicolumn{3}{c}{RefCOCO+}   & \multirow{2}{*}[-3pt]{Methods}  &\multicolumn{3}{c}{RefCOCO+}     \\ \cmidrule(l){2-4} \cmidrule(l){6-8} & $\text{val}^{d}$ & $\text{testA}^{d}$ & $\text{testB}^{d}$ & & $\text{val}^{d}$ & $\text{testA}^{d}$ & $\text{testB}^{d}$ \\ \midrule [\heavyrulewidth]

$\text{UNITER}_\text{BASE}$ \cite{chen2019UNITER}   & 71.5 & 77.0 & 60.1 
& $\text{VL-BERT}_\text{BASE}$ \cite{Su2019vlbert} & 70.7 & 76.8  & 60.3  \\ 

\ w/ DiM & \bf 72.7 & \bf 78.6 & \bf 62.1 
& \ w/ DiM & \bf 73.2 & \bf 78.9 & \bf 63.2  \\ \midrule 

ViLBERT \cite{lu2019vilbert} & 72.4  &78.3 &62.5    
& ESABERT & 73.4 & 79.3 & 63.7 \\

\ w/ DiM & \bf 74.3 & \bf 80.1 & \bf64.7 
& \ w/ DiM (DiMBERT) & \bf 75.6 & \bf 83.2 & \bf66.9 \\
\bottomrule
\end{tabular}

\end{table}

\subsection{Effect of DiM Module}
\label{sec:DiM}
In this subsection, we evaluate the effectiveness of the proposed Disentangled Multimodal-Attention (DiM) mechanism. We compare DiM with Entangled Self-Attention (ESA) which applies the same set of attention matrices to sentences, RoIs and concepts. As shown in Table~\ref{tab:effect_DMA}, the DiMBERT equipped DiM outperforms ESABERT equipped with ESA on both image caption and referring expressions tasks. Specifically, the DiM promotes the performance of DiMBERT from 73.4\% to 75.6\% and from 119.3 to 123.4 in terms of accuracies for RefCOCO+ and CIDEr for MSCOCO respectively.

This performance increase may attribute to disentangled attention. DiM explicitly use different attention matrices to model the visual and textual modalities, enabling better usage of the pre-trained BERT. As to the Entangled Self-Attention (ESA), the same set of attention matrices are used to model inner-vision, inner-language and mutual vision-language relations. Such multiple target optimization could affect language model's capability of the per-trained BERT.

DiM can also be easily integrated into existing pre-trained models. In this section, we further equip  $\text{UNITER}_\text{BASE}$ \cite{chen2019UNITER}, $\text{VL-BERT}_\text{BASE}$ \cite{Su2019vlbert} and ViLBERT \cite{lu2019vilbert} with DiM. Specifically, we pre-train these models on the Conceptual Captions dataset and evaluate the accuracy on the referring expression task. As shown in Table~\ref{tab:DiM}, DiM can successfully boost all baselines, with the most significant improvement up to relatively 4\%, 5\% and 5\% for $\text{val}^{d}$, $\text{testA}^{d}$ and $\text{testB}^{d}$, respectively.
The significant improvements demonstrate the effectiveness and universality of the proposed DiM module.

\begin{table}[t]
    \footnotesize
    \caption{Impact of visual concepts (ViCo). We further list the breakdown of SPICE F-scores \cite{anderson2016spice}, for a better understanding of the contribution of visual concepts.}
    \label{tab:effect_attribtues}
    \begin{tabular}{@{}l c c c c c c c c c@{}}
        \toprule
        \multirow{2}{*}[-3pt]{Methods} &RefCOCO+ &\multirow{2}{*}[-3pt]{B-4} &\multirow{2}{*}[-3pt]{M} &\multirow{2}{*}[-3pt]{C}  &\multicolumn{5}{c}{S} \\
        \cmidrule(lr){2-2} \cmidrule(l){6-10} & $\text{val}^{d}$ & & & & All  & Objects  & Attributes & Relations & Color \\ \midrule
        
        Base & \bf 68.5 & \bf  35.5 &\bf   28.0 & \bf  113.7 & \bf 20.8 &\bf 37.9 & \bf 10.2  & \bf 6.2 & \bf 10.2  \\ 
        
        \ w/o ViCo & 66.6  & 35.4 & 27.9 & 112.9 & 20.4 &37.3 & 9.3 &5.8 & 9.5 \\ 
        \midrule
        
        DiMBERT  & \bf 75.6 & \bf 38.1 & \bf 28.7  & \bf 123.4   & \bf 22.0 & \bf 40.0 & \bf 11.1 &\bf 7.3 & \bf11.5 \\ 
        
        \ w/o ViCo & 74.3 & 37.9 &  28.5  &120.2   &   21.5 & 39.4 & 9.9 &5.7 & 10.4   \\
        \bottomrule
        \end{tabular}

\end{table}

\begin{table}[t]
\footnotesize
\caption{Analysis about the effect of the number of visual concepts. All variants are conducted on the base model. We also report the performance of visual concept extractor in terms of the Precision, Recall and F1 scores. As we can see, when the number of visual concepts $M$ is 20, the visual concept extractor and the base model get the highest F1 score and highest performance.}
\label{tab:res_concept}
\begin{tabular}{@{}c c c c c c c c c c@{}}
\toprule
\multirow{2}{*}[-3pt]{Number} & \multirow{2}{*}[-3pt]{\begin{tabular}[c]{@{}c@{}}  Precision \\ (\%) \end{tabular} } & \multirow{2}{*}[-3pt]{\begin{tabular}[c]{@{}c@{}}  Recall \\ (\%) \end{tabular} } & \multirow{2}{*}[-3pt]{\begin{tabular}[c]{@{}c@{}}  F1 \\ (\%) \end{tabular} } & \multicolumn{1}{c}{RefCOCO+} &  \multicolumn{4}{c}{MSCOCO}     \\ \cmidrule(lr){5-5} \cmidrule(l){6-9}  & & & & $\text{val}^{d}$  & B-4  & M  & C & S     \\ \midrule [\heavyrulewidth]
$M$=0  & - & - & - & 67.4 & 35.0 & 27.7 &112.7 &20.4 \\
$M$=10 & \bf 72.6 & 29.8 & 44.3 & 68.1 & 35.3 & 27.9 &113.4 &20.6  \\
$M$=20 & 52.9 & 45.1  & \bf 49.5 & \bf68.3 & \bf35.5 & \bf28.0 & \bf113.7 & \bf20.8 \\
$M$=30 & 42.2 & 54.2 & 47.4 & 67.9 & 35.1 & 27.8 &113.5 &20.6   \\
$M$=40 &35.1  & \bf 59.9 & 43.8 & 67.7 & 35.2 & 27.8 &113.0 &20.5  \\
\bottomrule
\end{tabular}

\end{table}

\subsection{Effect of Visual Concepts}
\begin{figure*}[t]
\includegraphics[width=1\linewidth]{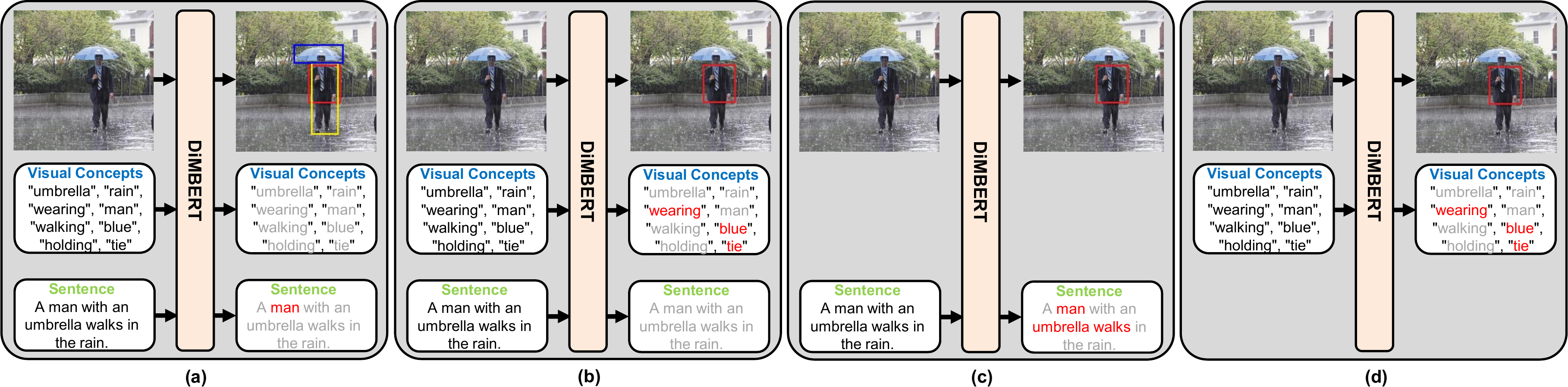}
\caption{Average attention weights of all heads in the last Disentangled Multimodal-Attention layer of DiMBERT. Please view in color. We show the alignments of a typical sentence word, i.e., \textit{man}, with top-3 image regions, i.e., text-to-RoI attention, in (a), and the the alignments of a typical image region, i.e., the \textbf{\color{red} Red} region, with top-3 semantic words, i.e., RoI-to-text attention, in (b), (c) and (d) . Specifically, (a) and (b) show that DiMBERT can provide visual-referred sentence representations and semantic-grounded image representations, respectively; (c) shows that the model is insufficient in providing suitable semantic information for regions without visual concepts; (d) shows that DiMBERT still works well even without input sentence.}
\label{fig:vis}
\end{figure*}

We conduct some experiments to investigate the contribution of visual concepts (ViCo) in our model.
Table~\ref{tab:effect_attribtues} shows that the visual concepts promote the baselines over all metrics, especially in \textit{Attributes} and \textit{Color}. The reason is that the visual concepts contain more \textit{attribute} words and \textit{color} words than the sentence. The abundant visual semantics in visual concepts can greatly enrich semantic representations of image regions. It can be confirmed in Figure~\ref{fig:vis} (c), which illustrates the image representations refined by DiMBERT w/o ViCo. Compared with the image representations refined by DiMBERT (Figure~\ref{fig:vis} (b)), it is obvious that the DiMBERT w/o ViCo is insufficient in providing suitable semantic information for image regions.
It is worth noticing that given the absence of input sentence, most existing models won't work well, but our model can still refine the image representations. The Figure~\ref{fig:vis} (d) shows the semantic-grounded image representations \cite{liu2019MIA} refined by our DiMBERT without input sentence.

Table~\ref{tab:res_concept} shows that when the number of visual concepts $M$ is 20, the visual concept extractor and the model get the highest F1 score and highest performance, respectively, which is the reason why the value of $M$ is set to 20 in our DiMBERT.
For other variants, we speculate that when $M$ is set to small values (lower recall score), the model will suffer from the inadequacy of information. When $M$ is set to large values (lower precision score), the module will introduce more noisy information.

\subsection{Visualization of DiMBERT}
To evaluate the effectiveness of DiMBERT, we visualize the alignments between visual regions and semantic words according to the attention weights in the last Disentangled Multimodal-Attention layer.
Figure~\ref{fig:vis} (a) and (b) show the examples of vision-language representations learned by our DiMBERT.
As we can see, the model provides visual references for the input sentence, e.g., the sentence word \textit{man} is aligned to the \textbf{\color{blue} blue}, \textbf{\color{yellow} yellow} and \textbf{\color{red} red} regions.
The visual-referred sentence representations play an important role in understanding the sentence correctly, because the visual references help alleviate semantic ambiguity, e.g., the word \textit{bank} can either refer to a financial organization or the side of a river, and the word \textit{mouse} can either refer to a mammal or an electronic device.
Similarly, the model can provide a clearer semantic information for image regions, e.g., the \textbf{\color{red} red} region is aligned to the words \textit{wearing}, \textit{blue} and \textit{tie}. Thus, the original image representations are refined to semantic-grounded image representations \cite{liu2019MIA}. 
Note that there is no suitable semantic word in the sentence to align with the \textbf{\color{red} red} region, thus DiMBERT does not assign too much attention weights for any sentence word, which also indicates the effectiveness of our model when it comes to insufficient sentence words.

The refined semantic-grounded image representations and visual-referred sentence representations are beneficial for understanding both image and sentence, providing a solid bias for vision-and-language tasks.

\begin{figure*}[t]
\includegraphics[width=1\linewidth]{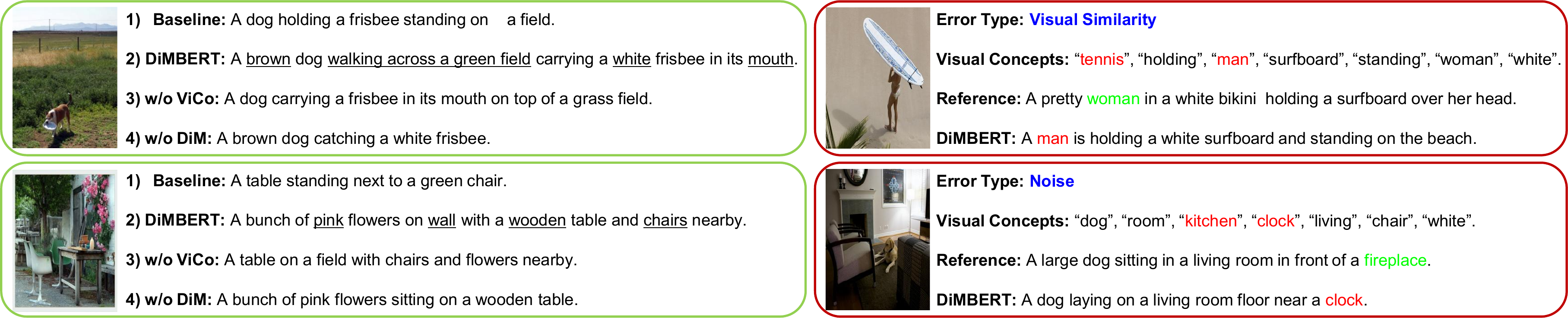}
\caption{Examples of the generated captions. The left plot and right plot show the correct examples and the error analysis of our approach, respectively. The ViCo and DiM represent the Visual Concepts and Disentangled Multimodal-Attention, respectively. The color \textbf{\color{green} Green} denotes desirable results, while \textbf{\color{red} Red} denotes unfavorable results.}
\label{fig:example}
\end{figure*}
\subsection{Examples and Bad Case Analysis}

In the left plot of Figure~\ref{fig:example}, we list some intuitive examples on MSCOCO image captioning task to find the the actual improvement.
Compared the $1^{\rm st}$ and $2^{\rm nd}$ lines, we find that the pre-training procedure helps the base model to generate more complete and accurate captions.
Compared the $2^{\rm nd}$, $3^{\rm rd}$ and $4^{\rm th}$ lines, we find that the visual concepts are bringing more details in colors and attributes, such as \textit{brown}, \textit{white}, \textit{pink} and \textit{wooden} than the ``w/o ViCo'' model, and the Disentangled Multimodal-Attention is bringing more comprehensive in objects, such as \textit{field}, \textit{mouth}, \textit{wall} and \textit{chairs} than the ``w/o DiM'' model, which corroborate the effectiveness of our approach.

We also analyze some bad cases to provide insights on how the DiMBERT may be improved. We find that there are mainly two types of errors, i.e., visual similarity and noise. In the right plot of Figure~\ref{fig:example}, we give some examples.
In the first example, our model dis-identify the \textit{woman} as a \textit{man} due to their visual similarity. However, humans can find that the person in the image wearing a bra.
In the second example, the DiMBERT mistakes the incorrect visual concept, i.e., \textit{clock}, for an appropriate one when it generates caption.
A more powerful ViCo extractor may be helpful in solving the problem, but it is unlikely to be completely avoided.

\section{Conclusions}

We present a visual concepts and proposed Disentangled Multimodal-Attention based pre-training model, i.e., DiMBERT, which is pre-trained on a large amount of image-sentence pairs to learn vision-language grounded representations. The pre-trained DiMBERT can be fine-tuned on various of vision-and-language tasks, including generation tasks and classification tasks. Extensive experiments and systematic analysis validate our motivations and corroborate the effectiveness and the universality of our Disentangled Multimodal-Attention and DiMBERT, where the latter sets new state-of-the-arts on three downstream tasks (over four datasets). 

\section*{Acknowledgments}
This work is partly supported by National Key R\&D Program of China (2020AAA0105200), Beijing Academy of Artificial Intelligence (BAAI), 
the IER foundation (No. HT-JD-CXY-201904) and Shenzhen Municipal Development and Reform Commission (Disciplinary Development Program for Data Science and Intelligent Computing). 
Special acknowledgements are given to Aoto-PKUSZ Joint Lab for its support.
We thank all the anonymous reviewers for their constructive comments and suggestions. Xu Sun and Yuexian Zou are the corresponding authors of this paper.

\bibliographystyle{ACM-Reference-Format}
\bibliography{sample-base}

\end{document}